\documentclass[letterpaper, 10 pt, conference]{IEEEtran}
 \def\IEEEtitletopspace{18pt}
\IEEEoverridecommandlockouts
\usepackage{cite}
\usepackage{amsmath,amssymb,amsfonts}
\usepackage{algorithm,algorithmic}
\usepackage{subfigure}
\usepackage[pdftex]{graphicx}
\usepackage{textcomp}
\usepackage{color}
\usepackage{xcolor}
\usepackage{pgfplots}
\pgfplotsset{compat=1.16}
\usepackage{tikz}
\usepackage{bm}
\usepackage[pagebackref=true]{hyperref} 

\newcommand\hlGreen[1]{%
  \bgroup
  \hskip0pt\color{green!80!black}%
  #1%
  \egroup
}
\newcommand\hlBlue[1]{%
  \bgroup
  \hskip0pt\color{blue!80!black}%
  #1%
  \egroup
}
\newcommand\hlPink[1]{%
  \bgroup
  \hskip0pt\color{magenta}%
  #1%
  \egroup
}

\newcommand\myeq{\mkern1.5mu{=}\mkern1.5mu}
\newcommand\myminus{\mkern1.5mu{-}\mkern1.5mu}
\newcommand\myless{\mkern1.5mu{<}\mkern1.5mu}

\DeclareMathOperator*{\argmin}{argmin}
\DeclareMathOperator*{\argmax}{argmax}
\newcommand{\etal}{\MakeLowercase{\textit{et al.}}}

\usepackage{color,soulutf8}
\usepackage{multirow}
\def\BibTeX{{\rm B\kern-.05em{\sc i\kern-.025em b}\kern-.08em
    T\kern-.1667em\lower.7ex\hbox{E}\kern-.125emX}}
\bstctlcite{IEEEexample:BSTcontrol}

\makeatletter
\newcommand{\linebreakand}{%
  \end{@IEEEauthorhalign}
  \hfill\mbox{}\par
  \mbox{}\hfill\begin{@IEEEauthorhalign}
}
\makeatother

\begin{document}

\title{Optimizing Demonstrated Robot Manipulation Skills for Temporal Logic Constraints
}

\author{\IEEEauthorblockN{
Akshay Dhonthi\IEEEauthorrefmark{1}\IEEEauthorrefmark{2}\textsuperscript{\textsection},
Philipp Schillinger\IEEEauthorrefmark{1},
Leonel Rozo\IEEEauthorrefmark{1} and
Daniele Nardi\IEEEauthorrefmark{3}}
\IEEEauthorblockA{\IEEEauthorrefmark{1}
Bosch Center for AI, Renningen, Germany \\
\IEEEauthorblockA{\IEEEauthorrefmark{2}
Formal Methods and Tools, University of Twente, Enschede, Netherlands\\ }
\IEEEauthorblockA{\IEEEauthorrefmark{3}
Department of AI and Robotics, Sapienza University, Rome, Italy\\ }
Email: \IEEEauthorrefmark{2}a.dhonthirameshbabu@utwente.nl,
\IEEEauthorrefmark{1}\{philipp.schillinger, leonel.rozo\}@de.bosch.com, \\ \IEEEauthorrefmark{3}nardi@diag.uniroma1.it}}

\maketitle

\begingroup\renewcommand\thefootnote{\textsection}
\footnotetext{This work was carried out at Bosch Center for AI and Sapienza University.}
\endgroup

\begin{abstract}
For performing robotic manipulation tasks, the core problem is determining suitable trajectories that fulfill the task requirements.
Various approaches to compute such trajectories exist, being learning and optimization the main driving techniques.
Our work builds on the learning-from-demonstration (LfD) paradigm, where an expert demonstrates motions, and the robot learns to imitate them.
However, expert demonstrations are not sufficient to capture all sorts of task specifications, such as the timing to grasp an object.
In this paper, we propose a new method that considers formal task specifications within LfD skills.
Precisely, we leverage Signal Temporal Logic (STL), an expressive form of temporal properties of systems, to formulate task specifications and use black-box optimization (BBO) to adapt an LfD skill accordingly.
We demonstrate our approach in simulation and on a real industrial setting using several tasks that showcase how our approach addresses the LfD limitations using STL and BBO.
\end{abstract}

\section{Introduction}
Learning from demonstration (LfD) is a paradigm that uses expert demonstrations to derive robot control policies~\cite{atkeson1997robot}.
In manipulation tasks employing torque-controlled industrial robots, kinesthetic teaching is often exploited, where an expert operator physically moves the robotic arm to generate demonstrations of a task~\cite{ravichandar2020recent}.
However, these demonstrations alone may not be sufficient for learning tasks that involve contacts or high accuracy and reliability~\cite{zhu2018robot}.

In industrial settings, there are several limitations while generating demonstrations:
First, it is difficult for the expert to show accurate time constraints in the task execution.
For instance, reaching or avoiding a specific region during a given time interval.
Second, minor additions to the task require additional full demonstrations to subsequently update the learning model.
These additions may be spatial (e.g. choosing an alternative path) or temporal (e.g. completing different parts of the task at different speeds or time intervals).
And third, the demonstrations do not always match the desired task performance, and therefore there may be room for improvement in task reliability, accuracy, or execution time.

Some of the aforementioned challenges may be addressed by learning reward functions from expert examples via inverse reinforcement learning~\cite{Adams22:SurveyIRL}.
However, the reward function learning heavily depends on expert data, which means that when new task requirements arise, the human often needs to provide additional demonstrations to trigger a new training process of the model or the reward function.
Also, defining the specific structure of reward functions is not trivial.


\begin{figure*}[ht]
    \centering
    \begin{tikzpicture}
        \node (image) at (-4.6,1.35) {\includegraphics[width=2.5cm]{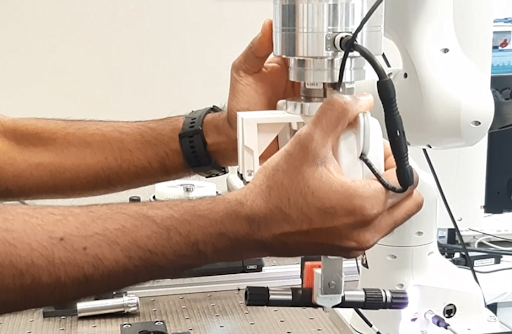}};
        \node (image) at (-4.6,-1.65) {\includegraphics[width=3.1cm]{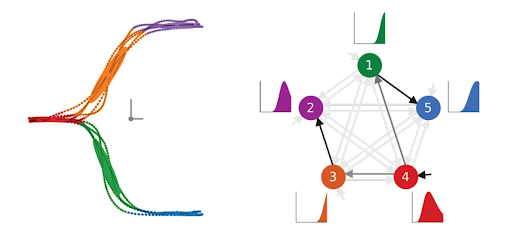}};
        \node (image) at (6.5, -1.50) {\includegraphics[width=2.25cm]{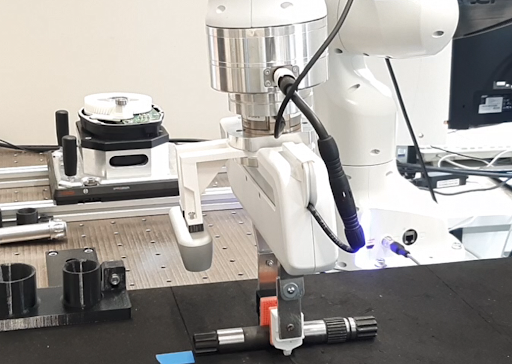}};
        \node (image) at (3.2, -1.25) {\includegraphics[width=2.0cm]{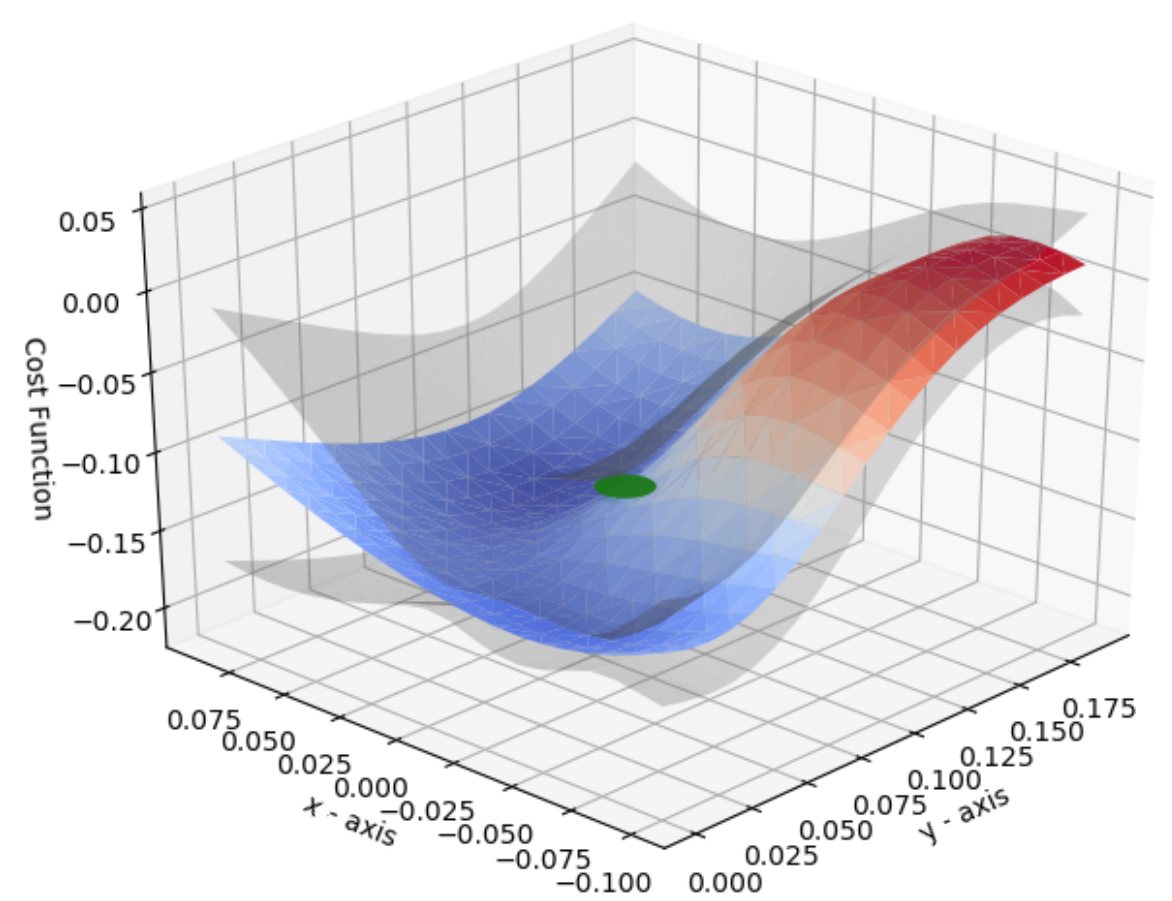}};
        \node (image) at (3.0, 1.30) {\includegraphics[width=2.4cm]{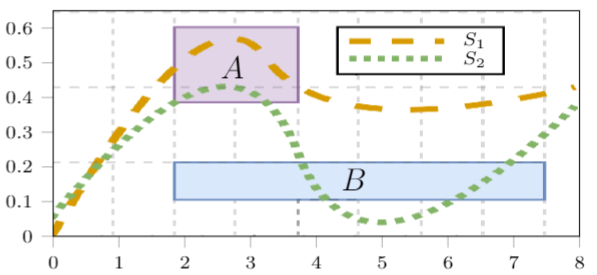}};

        \draw (-4.6, 1.55) node[rectangle, minimum height=2.25cm, minimum width=4.0cm, draw] (demo) {};
        \node at (-4.70, 2.45){\normalsize \textbf{Human Demonstrations}};

        \draw (-4.6, -1.25) node[rectangle, minimum height=2.25cm, minimum width=4.5cm, draw] (model) {};
        \node at (-5.5, -0.40){\normalsize \textbf{Model Learning}};
        \node at (-4.60, -0.85){\normalsize \underline{Hidden Semi-Markov Model}};

        \draw (1.3, 1.55) node[rectangle, minimum height=2.25cm, minimum width=6cm, draw] (task) {};
        \draw (1.3, -1.25) node[rectangle, minimum height=2.25cm, minimum width=6cm, draw] (bbo) {};
        \draw (6.5, -1.25) node[rectangle, minimum height=2.25cm, minimum width=3.1cm, draw] (skill) {};
        
        \draw[-latex] (demo.south) -- (model.north);
        \draw[-latex] (model.east) -- (bbo.west);
        \draw[-latex] (task.south) -- (bbo.north);
        \draw[-latex] (bbo.east) -- (skill.west);
        \draw[-latex] (skill.west) -- (bbo.east);
        
        \node at (-0.2, 2.45){\normalsize \textbf{Task Constraints}};
        \node at (0.10, 1.75){\normalsize \underline{Signal Temporal Logic}};
        \node at (0.10, 1.30){\normalsize Ex: Visit B in 5 sec};
        \node at (0.10, 0.85){\normalsize $\varphi = \mathbf{F}_{[0,5]}\varphi_{B} $};

        \node at (0.2, -0.40){\normalsize \textbf{Bayesian Optimization}};
        \node at (0.30, -1.20){\normalsize Refine model parameters};
        \node at (0.30, -1.55){\normalsize until task satisfaction};

        \node at (6.5, -0.40){\normalsize \textbf{Skill Reproduction}};

        \end{tikzpicture}
    \caption{Illustration of the proposed method. We collect several demonstrations of robotic manipulation skills, which are then used to learn a TP-HSMM model. This model encapsulates the observed spatial and temporal patterns of the demonstrations. We then define new task constraints based on STL specifications. Finally, we leverage Bayesian optimization to update the TP-HSMM model parameters in order to fulfill the new STL-based task requirements.}
    \label{fig: Illustrations}
    \vspace{-0.3cm}
\end{figure*}
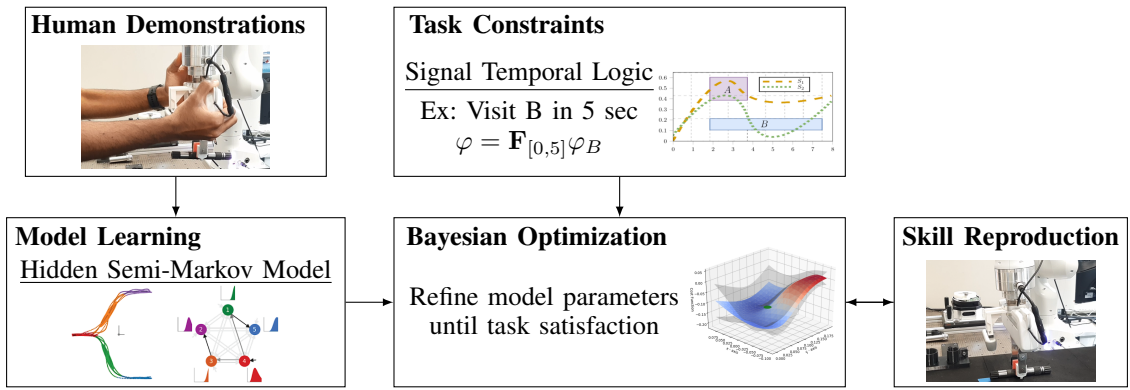

Due to these difficulties, we tackle this problem from an optimization perspective, as illustrated in Fig.~\ref{fig: Illustrations}.
In our method, when new task requirements arise, we use a black-box optimization strategy to refine the LfD model parameters.
This optimization exploits an objective function that captures the desired spatial and temporal constraints of the new task using Signal Temporal Logic (STL)~\cite{mehdipour2019arithmetic}.
STL is a formalism capable of representing high-level task objectives as logical semantics, making their design more user-friendly.
STL provides a real-valued function called \emph{robustness degree}, generated from a logical specification, which makes the evaluation of task satisfaction appealing for black-box optimizers.

There have been several works on leveraging STL on robotic applications~\cite{aksaray2016q, araki2019learning, xu2019transfer, li2017reinforce}.
Aksarya \etal~\cite{aksaray2016q} proposed an optimal policy learning scheme for satisfying STL specifications.
Although this method does not fully fall under our LfD-based approach, it is worth highlighting that STL specifications can be quantified and used as reward functions for learning a robot policy.
Additionally, unlike~\cite{aksaray2016q} where the agent acts in a discrete environment, our approach focuses on continuous signals. 
Innes \etal~\cite{innes2020elaborating} used neural networks to learn the expert demonstrations.
For these networks, \emph{Linear Temporal Logic} (LTL) specifications were transformed into a differentiable loss function.
In contrast, our approach uses STL and is therefore more general because STL captures continuous propositions and robustness, as well as time intervals.

The closest approach to our work is~\cite{puranic2021learning}, where a quantified STL is used to rank the quality of demonstrations as a function of robustness.
Then, their approach learns a reward function from which an optimal policy is derived via reinforcement learning.
This technique explores trajectories around the demonstrations as possible states leading to high rewards.
Our approach differs from this technique in two ways:
First, we exploit expert demonstrations directly to learn an LfD model without accounting for STL during this learning stage.
Second, we consider scenarios where new task requirements, previously unseen, may be added to existing skills.

In summary, our contributions are threefold:
(1) We leverage STL specifications to design explicit task objectives on top of previously-learned robotic skills;
(2) We demonstrate that Bayesian optimization for these STL-based specifications overcomes the issue of suboptimal demonstrations and allows the operator to include new task requirements in the skill model;
(3) We validate our approach in two experiments that exploit STL rewards to achieve task objectives given by a human operator, implemented in simulation and on a real industrial setup.

\section{Background}
This section provides a brief introduction to STL and computation of traditional robustness degree using an example.
Further, we introduce the optimization technique and the LfD model used in this work.

\subsection{Signal Temporal Logic (STL)} \label{Sec: STL}
Formally, an STL specification can be understood as follows: 
Consider a discrete time sequence\footnote{
We use subscripts to denote sequences of data. i.e. $t_{0:k}=\{t_0, \cdots ,t_k\}$} $t_{0:k} \in \mathbb{R}^k$.
The STL formula $\varphi$ is defined using the predicate $\mu$ 
(an atomic proposition, i.e. a point-wise constraint using $>$ or $<$ operators)
that is of the form $f(x(t_{0:k}))$, where $x(t_k)$ is the state of the signal\footnote{The signal is in discrete form with $k$ steps and it is referred to as $x(t)$ throughout the paper for simplicity.}
(e.g. a robot trajectory, joint velocity, joint torques, etc.)
at time $t_k$.
Moreover, the predicate function $f:\mathbb{R}^k \rightarrow \mathbb{R}$ maps each time point to a real-value~\cite{aksaray2016q, mehdipour2019arithmetic}.
Then, the STL syntax is defined as,
\begin{equation}
    \begin{aligned}
        \varphi \;  := & \; \mu \; | \; \neg\varphi \; | \; \varphi_1\land\varphi_2 \; | \; \varphi_1\lor\varphi_2 \; \\
        & | \; \mathbf{G}_{I}\varphi \; | \; \mathbf{F}_{I}\varphi \; | \; \varphi_1 \mathbf{U}_{I}\varphi_2 ,
    \end{aligned}
    \label{Eqn: STL}
\end{equation}
where $I = [a, b]$ is the non-empty set of all $t \in t_{0:k}$ such that $a \leq t \leq b$.
The operators $\neg, \land, \lor$ refer to Boolean \emph{negation}, \emph{conjunction} and \emph{disjunction}, respectively.
The temporal operators $\mathbf{G}, \mathbf{F}, \mathbf{U}$ represent \emph{globally}, \emph{eventually} and \emph{until} statements, respectively.
The satisfaction of $\mu$ is $\textsc{True} \; (\top)$ if the predicate is satisfied and $\textsc{False} \; (\bot)$ otherwise.
The globally $\mathbf{G}_{[a,b]}$ operator states that $\varphi$ must be true at all times in $[a, b]$, and the eventually $\mathbf{F}_{[a,b]}$ operator states that $\varphi$ must be true at some time point in $[a, b]$.
Similarly, the until $\mathbf{U}_{[a,b]}$ operator states that $\varphi_1$ must be true until $\varphi_2$ eventually becomes true in the time interval $[a, b]$.

\begin{figure}[t]
    \centering
            \begin{tikzpicture}
            \begin{axis}[
                xlabel={Time (sec)},ylabel={Signal},
                xtick distance=1,ytick distance=0.1,
                xtick pos=bottom, ytick pos=left,
                grid=both,
                label style={font=\scriptsize},
                tick label style={font=\scriptsize, /pgf/number format/fixed},
                axis on top, 
                width=6.75cm,
                height=2.75cm,
                scale only axis=true,
                enlargelimits=false,
                tick align=outside,
                ]
            \addplot graphics[xmin=0, xmax=8.50, ymin=0.0, ymax=0.65, includegraphics={width=7.5cm, trim={0.0cm 0.0cm 0.0cm 0.0cm},clip}] {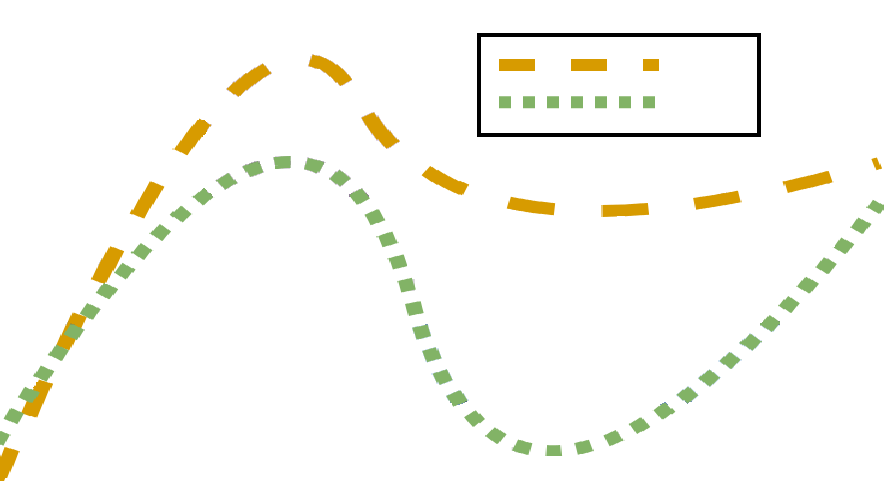};
            \end{axis}
            \draw (2.38, 2.13) node[rectangle, fill=purple, opacity=0.2, minimum height=0.84cm, minimum width=1.55cm, draw] (rectA) {};
            \draw (3.975, 0.615) node[rectangle, fill=blue, opacity=0.2, minimum height=0.43cm, minimum width=4.75cm, draw] (rectB) {};
            \node at (5.4,2.40){\scriptsize $S_1$};
            \node at (5.4,2.15){\scriptsize $S_2$};
            \node at (2.35,2.18){\large $A$};
            \node at (4.0,0.63){\large $B$};
        \end{tikzpicture}
    \vspace{-0.3cm}
    \caption{For the STL specification to visit region \textsc{A} and avoid region \textsc{B}, the signal $S_1$ satisfies it and the signal $S_2$ violates it.}
    \label{fig:STL Example}
    \vspace{-0.3cm}
\end{figure}
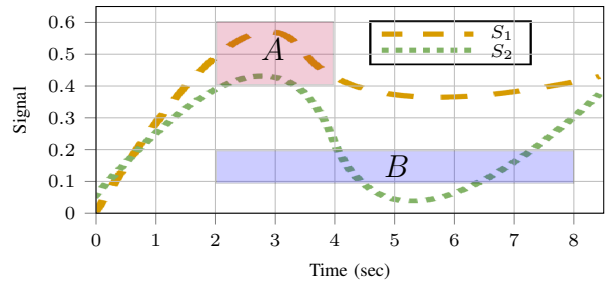

As an example, let us consider the two one-dimensional signals shown in Fig.~\ref{fig:STL Example}.
Also, consider the requirement: ``The signal should pass through region \textsc{A} during $2$ to $4$ seconds and avoid region \textsc{B} during $2$ to $8$ seconds". 
This can be translated into an STL specification as follows:
\begin{equation}
\begin{aligned}
    \varphi =& \mathbf{F}_{[2,4]}\varphi_{A} \; \land \; \mathbf{G}_{[2,8]} \neg\varphi_{B},
\end{aligned}
\label{Eqn: STL Example}
\end{equation}
where $\varphi_A$ is given by the bounds $0.4 < S_i(t) < 0.6$. The proposition for this is $f_A(S_i(t)) := 0.1 - |S_i(t) - 0.5|$. Also, $\varphi_B$ is defined accordingly.
Figure~\ref{fig:STL Example} shows how the signal $S_1$ satisfies the specification given by~\eqref{Eqn: STL Example} whereas $S_2$ fails due to the violated requirement of avoiding region \textsc{B}.

\subsection{Robustness Degree} \label{Sec: RobMetric}
The robustness degree, denoted as $r(\varphi, x, t) \in \mathbb{R}$, is the quantitative semantics of the STL formula $\varphi$ that measures ``how well" the signal $x$ is fulfilled at time $t$.
The classical way of defining these semantics is via \emph{space robustness}~\cite{belta2019formal}. 
This measure is positive if and only if the signal satisfies the specification (i.e. \emph{soundness} property). The closer the robustness is to zero, the smaller the required changes of signal values are to change the truth value.
Formally, space robustness and its corresponding operators are defined as follows
\begin{equation}
    \begin{aligned}
        r(\mu, x, t) =& f(x(t)) ,\\
        r(\neg\varphi, x, t) =& -r(\mu, x, t) ,\\
        r(\varphi_1\land\varphi_2, x, t) =& \min (r(\varphi_1, x, t), r(\varphi_2, x, t)) ,\\
        r(\varphi_1\lor\varphi_2, x, t) =& \max (r(\varphi_1, x, t), r(\varphi_2, x, t)) ,\\
        r(\mathbf{G}_{[a,b]}\varphi, x, t) =& \min_{t_k \in [t+a, t+b]} (r(\varphi, x, t_k)) ,\\
        r(\mathbf{F}_{[a,b]}\varphi, x, t) =& \max_{t_k \in [t+a, t+b]} (r(\varphi, x, t_k)) .\\
    \end{aligned}
    \label{Eqn: SpaceRobustness}
\end{equation}

Concerning our example depicted in Fig.~\ref{fig:STL Example}, the robustness value is $0.1$ for the signal $S_1$ and $-0.05$ for the signal $S_2$.
The signal $S_2$ gets a negative reward since it fails to satisfy the global condition to avoid region \textsc{B}.

\subsection{Black-box Optimization (BBO)}
Black-box optimizers are widely used nowadays in several fields of machine learning~\cite{alarie2021two, pmlrv133turner21a}.
BBO is a sample-efficient technique to find optimal system parameters that maximize a task-specific objective function~\cite{watanabe2014black}.
In a typical BBO setting, the objective and constraint functions are only accessible through (possibly noisy) output values, and therefore gradient-based approaches are infeasible. 
For this work, we use \emph{Bayesian Optimization} (BO), which seeks an optimal set of parameters,
\begin{equation}
    \bm{\theta}^* = \argmin_{\bm{\theta}\in \bm{\Theta}} g(\bm{\theta}) .
\label{Eqn: BO}
\end{equation}

BO is a class of sequential search algorithms for optimizing black-box functions~\cite{brochu2010tutorial}. 
BO finds a global minimizer $\bm{\theta}^*$ as in~\eqref{Eqn: BO}, where $g$ is an unknown objective function defined over a domain of interest $\bm{\Theta}$.
The function $g$ is not observed directly, as only noisy output values are available.
Using this noisy information about the objective function, BO selects a query point $\bm{\theta}\in\bm{\Theta}$ at which to evaluate $g$ at each optimization step. 
Such a selection process is carried out by optimizing an acquisition function, which resolves the explore-exploit trade-off.
Several forms of acquisition function exist in the literature, and we use \empty{probability of improvement} (PI) in this work~\cite{kushner1964new}.
PI is a greedy search algorithm that selects the most likely point to offer an improvement.

\subsection{Learning from Demonstration (LfD)} \label{Sec: LfD}
LfD may be seen as a robot programming strategy that uses human demonstrations to build a learning model that encapsulates the main motion patterns of a skill or task.
As our interest is on learning object-centric manipulation skills, we leverage \emph{Task-Parameterized Hidden Semi-Markov Models} (TP-HSMM)~\cite{kumar2018generalizing} which have been successfully used in elaborated industrial tasks~\cite{Rozo20:Sequencing}.
TP-HSMM is an object-centric model that can be easily used to sequence several skills to accomplish complex assembly tasks~\cite{Rozo20:Sequencing}.
This parametric model encapsulates skill-specific spatial and temporal patterns from human demonstrations. 
Such encoding features are particularly useful in our setting due to their close relationship to the spatial and temporal requirements specified via STL.

Specifically, a TP-HSMM model learns a joint probability density function of the demonstrations using an object-centric formulation of hidden semi-Markov models (HSMM)~\cite{yu2010hidden}.
The underlying process of an HSMM is a Markov chain with one extension:
Each model state has variable duration, affecting how the system transits between states.
The TP-HSMM parameters are estimated by following a modified version of the Expectation-Maximization algorithm~\cite{kumar2018generalizing}.
Then, a linear quadratic tracker is used to synthesize a smooth trajectory following the main spatial and temporal patterns, encapsulated in the TP-HSMM model, as proposed in~\cite{kumar2018generalizing,Rozo20:Sequencing}. 
Note that the task-parameterized formulation allows the robot to adapt its motion to the pose changes in the manipulated objects.

Here we provide a short description of the TP-HSMM parameters relevant to the subsequent STL-based optimization process.
A set of $K$ states characterizes a TP-HSMM. They are usually represented by normal distributions $\mathcal{N}(\bm{\mu}_{k}, \bm{\Sigma}_{k})$ with $k=1 \ldots K$, which model the observation probabilities of the demonstrations on a set of coordinate systems (a.k.a. task parameters)\footnote{For the sake of simplicity, we assume that the demonstrations correspond to a single task-parameter setting.}.
The \emph{transition probability matrix} $\textbf{A} \in \mathbb{R}^{K \times K}$ with $A_{i, j} \triangleq P(z_t = j | z_{t-1} = i)$, defines the probability of transiting from state $i$ to state $j$ and encapsulates sequential information of the skill.
Finally, the \emph{duration probability} is represented by a normal distribution
$\mathcal{N}(s|\mu_j^S, \Sigma_j^S)$, and indicates how long the robot stays in the model state $j$, therefore encapsulating temporal information of the skill.
The spatial and temporal patterns of skill are then fully described by the duration probability, the transition matrix, and the observation probabilities, as detailed in~\cite{yu2010hidden}.

\section{Methodology}\label{Sec: Methodology}
The core idea of our approach is to integrate LfD models, BO, and STL, to handle new spatial or temporal task requirements for fast skill adaptation.
One of the first challenges in this context is to define the objective function and parameter space of interest.
On the one hand, the objective function (reward function) is designed from the robustness degree $r(\varphi,x,t)$ computed via the STL specification $\varphi$, introduced in Sec.~\ref{Sec: STL}.
On the other hand, the parameter space can be defined as a (sub)set of the skill model parameters.
Importantly, we can exploit the variance information encoded by the TP-HSMM model to design the corresponding bounds for the set of parameters as in~\cite{Rozo19:LfDBO}.
We therefore assume that the optimal solution is safe as long as the variance-based bounds are not violated.
This section covers both challenges in detail and illustrates them through several examples.

Algorithm~\ref{Alg: Learning} shows the STL-based Bayesian optimization procedure.
The goal is to refine the model parameters $\bm{\delta}$ such that the new task requirements, defined via STL specifications, are satisfied.
To begin with, we set the optimization parameters $\bm{\delta}_i$ at random for the first $M$ iterations.
Using these initial parameters set, we evaluate the objective function during the corresponding runs on the real system.
Following $M$ evaluations, as indicated in line~\ref{alg: line: acquisition}, we find the next set of parameters by maximizing the acquisition function $u$ (in this case, \emph{PI}), which directs the search toward the optimum. 

\begin{algorithm}[t]
    \caption{STL-based Bayesian optimization of LfD skills }
    \begin{algorithmic}[1]
        \renewcommand{\algorithmicrequire}{\textbf{Input:}}
        \renewcommand{\algorithmicensure}{\textbf{Output:}}
        \REQUIRE Optimization iterations $N$, number of evaluations $M$ to build initial observations set, STL specification $\varphi$, and TP-HSMM parameters $\bm{\delta}_0 \in \mathbb{R}^P$, where $P$ is the total number of considered parameters.
        \ENSURE  Updated model parameters $\bm{\delta}^* \in \mathbb{R}^P$
        \FOR {$i = 1,2,\cdots N$}
            \IF {($i \leq M$)}
                \STATE Randomly set $\bm{\delta}_i$ under respective upper and lower bounds $\bm{\mathcal{B}}_{1:P} \in \mathbb{R}^{2P}$ \label{alg: line: randomInit}
            \ELSE
                \STATE Find $\bm{\delta}_i$ under bounds $\bm{\mathcal{B}}_{1:P}$ by optimizing the PI acquisition function using the GP information: $\bm{\delta}_i~=~\argmax_\delta u(\bm{\delta}|\mathcal{D}_{1:i-1})$ \label{alg: line: acquisition}
            \ENDIF
            \STATE Skill retrieval using new model parameters $\bm{\delta}_{i}$ and linear quadratic tracking as in~\cite{Rozo20:Sequencing}. \label{alg: line: retrieve}
            \STATE Execute the skill and record signals of interest.
            \label{alg: line: log}
            \STATE Calculate the robustness $r(\varphi,x,t)$ to quantify the STL specification $\varphi$ for signal $x$. \label{alg: line: robDegree}
            \STATE Augment the data $\mathcal{D}_{1:i} = \{\mathcal{D}_{1:i-1}, (\bm{\delta}_i, r(\varphi,x,t)) \}$ and update the GP model. \label{alg: line: update}
         \ENDFOR
        \RETURN The model parameters with maximum reward $r(\varphi,x,t)$ from the data $\mathcal{D}_{1:N}$ 
    \end{algorithmic} 
\label{Alg: Learning}
\end{algorithm}

At each evaluation step $i$ in line~\ref{alg: line: retrieve}, we obtain a new skill trajectory for the respective parameter set $\bm{\delta}_i$ using the retrieval technique discussed in Sec.~\ref{Sec: LfD}.
We run this trajectory and record the relevant multi-dimensional signal $\bm{x}(t)$ as a discretized time sequence corresponding to the specifications' propositions.
Further, line~\ref{alg: line: robDegree} uses the STL specification $\varphi$ and the logged signal to compute the robustness degree $r(\varphi,x,t)$ that acts as the objective function.
The recorded observations and the \emph{Gaussian Process} (GP) are updated accordingly before the next iteration.
We run the optimization process for $N$ iterations to produce an optimal trajectory that maximizes the objective function.
The following subsection describes a thorough approach for calculating the robustness degree.

\subsection{Objective Function}\label{Sec:LearnSTL}
We tackle the robustness degree computation here, particularly the technique \emph{New Robustness} \cite{varnai2020robustness}.
We outline the definition of operators of \emph{New Robustness} and refer to the original work in~\cite{varnai2020robustness} for further details.
This computation relies on the multi-dimensional signal $\bm{x}(t)$ logged at each iteration of Algorithm~\ref{Alg: Learning}, see line~\ref{alg: line: log}.
The signal captures relevant measures for the stated specification, such as the robot position in task space, joint torques, joint velocities, or contact forces.
The STL specification can then contain several predicates $\mu$ that evaluate this signal.
To recap Sec.~\ref{Sec: STL}, each predicate $\mu$ is of the form $f(x_p(t))$ where $f$ maps each point in the signal to a real value.

The robustness degree, \emph{New Robustness}~\cite{varnai2020robustness}, was
experimentally compared against alternative robustness formulations in our previous work~\cite{dhonthi2021study}.
We choose \emph{New Robustness} here because it performed best in finding optimal solutions with a faster convergence rate for the manipulation tasks studied in~\cite{dhonthi2021study}.
The computation of \emph{New Robustness} is based on the elementary definitions of the negation $\neg$ and conjunction $\land$ operators.
The paper~\cite{varnai2020robustness} introduced a structured definition of robustness degree with which it is possible to derive all the boolean and temporal operators of STL using just the $\neg$ and $\land$ operators.
Additionally, the $\neg$ operator can also be excluded from the definition by setting it to a negative value of $\land$.
Therefore, the computation of any STL operator boils down to the definition of the conjunction operator, which is
\begin{equation}
    \begin{aligned}
        (\varphi_1\land\cdots\land\varphi_m) := 
        \begin{cases} 
            \dfrac{\sum_i \rho_{\min}e^{\tilde{\rho}_i}e^{\nu\tilde{\rho}_i}}{\sum_i e^{\nu\tilde{\rho}_i}} & if \; \rho_{\min} < 0, \\
            \dfrac{\sum_i \rho_{i}e^{-\nu\tilde{\rho}_i}}{\sum_i e^{-\nu\tilde{\rho}_i}} & if \; \rho_{\min} > 0, \\
            0 & if \; \rho_{\min} = 0 ,
        \end{cases}
    \end{aligned}
    \label{Eqn: NewRob}
\end{equation}
with,
\begin{equation}
\begin{aligned}
    \rho_i \myeq r(\varphi_i, x, t), \
    \rho_{\min} \myeq \min(\rho_1\cdots\rho_m), \
    \tilde{\rho}_i \myeq \frac{\rho_i \myminus \rho_{\min}}{\rho_{\min}},
\end{aligned}
\end{equation}
where $\rho_{\min}$ is the quantified conjunction operator using traditional space robustness. The parameter $\nu > 0$ tends to traditional conjunction as $\nu \rightarrow \infty$.

This robustness formalism is suitable in the robot task definition for the following reasons:
It satisfies the soundness property; as for any specification, the robustness degree returns a positive value if and only if the signal satisfies $\varphi$ at time $t$.
It is achieved by preserving the sign in the definition of $\rho_{\min}$ (see~\cite{varnai2020robustnessproof} for the proof).
Thus, the output of \eqref{Eqn: NewRob} quantifies how much a task is satisfied or violated, which is beneficial in manipulation tasks where complete satisfaction is crucial.
This robustness degree also captures partial progress towards the goals and is beneficial in faster guiding towards task optimum.
These properties are beneficial for \emph{reward shaping} \cite{ng1999policy} to obtain indicative samples during the optimization procedure and for fitting the GP, although the relation between a signal $x$ and the skill model parameters $\bm{\delta}$ remains a black-box function.

We turn the attention to two examples with different STL specifications for illustrating the approach:
Consider a simple skill in which the robot end-effector visits three regions $L_{1:3}$ in Cartesian space.
We first learn a TP-HSMM model with $K$ components from human demonstrations.
The new task requirements that must be satisfied using our approach are 
\begin{enumerate}
    \item While satisfying the main task of visiting three regions $L_{1:3}$, the end-effector must pass through another region $L_4$ in the time interval $(8, 12)$ seconds and stay inside the region $L_2$ during $(12, 15)$ seconds.
    The STL specification for these new task requirements is given as
\begin{equation}
    \varphi_1 = \mathbf{F}\varphi_{L_1} \land \; \mathbf{G}_{[12,17]}\varphi_{L_2} \land \; \mathbf{F}\varphi_{L_3} \land \;  \mathbf{F}_{[8,12]}\varphi_{L_4} .\\
\label{Eqn: STL Spec 1}
\end{equation}

    \item The robot end-effector must visit regions $\{L_1, L_3\}$ at any time, avoid region $L_2$ during $(12, 17)$ seconds and visit a new region $L_4$ during the time interval $(8, 12)$ seconds. 
    The STL specification for these new task conditions is
\begin{equation}
    \varphi_2 = \mathbf{F}\varphi_{L_1} \land \; \mathbf{G}_{[12,17]} \neg\varphi_{L_2} \land \; \mathbf{F} \varphi_{L_3} \land \;  \mathbf{F}_{[8,12]}\varphi_{L_4} .\\
\label{Eqn: STL Spec 2}
\end{equation}

\end{enumerate}

The regions $L_i$ with $i \in \{1, 2, 3, 4\}$ are specified using Cartesian bounds as follows
\begin{equation}
\begin{aligned}
    \varphi_{L_i} \myeq & x_{i,lb} \myless x \myless x_{i,ub} \ \land \
    y_{i,lb} \myless y \myless y_{i,ub} \ \land \\
    & z_{i,lb} \myless z \myless z_{i,ub}.
\end{aligned}
\label{Eqn: STL Regions}
\end{equation}
Note that the required signals in~\eqref{Eqn: STL Regions}  correspond to the Cartesian coordinates $x, y, z$  of the robot end-effector.

Let us assume that we ran a trial of the experiment and recorded the aforementioned signals. 
We discretize them and, for simplicity, define the signals of $x, y, z$ coordinates as $x(t), y(t), z(t)$, respectively.
Each region has three propositions, one for each Cartesian bound.
For instance, the proposition of the $x$-coordinate (similarly for $y$ and $z$) for the regions $L_i$ is written as follows,
\begin{equation}
\begin{aligned}
f_{L_i, x}(x(t)) :=& \frac{x_{i,ub} - x_{i,lb}}{2} - \left| x(t) - \frac{x_{i,ub} + x_{i,lb}}{2} \right|. \\
\end{aligned}
\label{Eqn: RegionProp}
\end{equation}
At the end of these steps, we have several propositions separated by operators.
We now use~\eqref{Eqn: NewRob} to compute a real-value that describes how well a skill satisfied the task.
Using this, the GP is updated, and we get new model parameters for the next iteration.
The loop is continued until termination.

\subsection{Parameter Space}
The parameter space is defined as $\bm{\delta} = \{\bm{\mu}_{1:K}, \bm{\mu}_{1:K}^S, \bm{A}\}$, where $\bm{\mu}_{1:K}$ are the Gaussian components positions, $\bm{\mu}_{1:K}^S$ are the duration probabilities, and $\bm{A}$ is the transition matrix.
Next, we discuss the role of each of these parameters in detail.

\subsubsection{Component positions}
As described in Sec.~\ref{Sec: LfD}, each TP-HSMM state is represented by a Gaussian distribution $\mathcal{N}(\bm{\mu}, \bm{\Sigma})$.
Once a skill model is learned, these Gaussian components are used to retrieve a trajectory to be reproduced by the robot.
Therefore, we can modify the TP-HSMM states to reshape the robot trajectory. 
For instance, modifying the mean $\bm{\mu} \in \mathbb{R}^3$ will translate the Gaussian in Cartesian space.
Therefore, we adapt $\bm{\mu}$ at each iteration under certain bounds $\bm{\mathcal{B}}_\mu$ for each component.
Bounds define how far the components may move in any direction.
The maximum number of parameters considered for optimization is $K \times 3$ as each component $\bm{\mu}$ has three parameters to represent its position in 3D space.
Shifting the components is essential to address the spatial constraints of a task.
For instance, refining $\bm{\mu}$ is beneficial in the STL specification $\varphi_1$ where a new region $L_4$ has to be visited at a particular time interval.

\subsubsection{Duration probabilities}
Each TP-HSMM state is also assigned a duration probability represented by a normal distribution $\mathcal{N}(s|\mu^S, \Sigma^S)$, which defines how long to stay in that model state.
Raising or lowering the value of the mean $\mu^S \in \mathbb{R}$ increases or decreases the time spent in the respective component, thus affecting the task timing. 
We adapt $\mu^S$ under the bounds $\mathcal{B}_{\mu^S} = [-\sigma^S, \sigma^S]$, where $\sigma^S$ is the corresponding standard deviation of the duration probability.
The maximum number of parameters is $K$ when considering to optimize duration probabilities of all components.
Modifying duration probabilities allows us to optimize the temporal constraints of a task.
For instance, adapting $\mu^S$ is beneficial in $\varphi_1$, where the condition to globally stay in the region $L_2$ for the time interval $(12, 17)$ is possible by reducing the duration of all the components except the one associated with $L_2$.

\subsubsection{Transition matrix}
The matrix $\bm{A}$ defines the probability of transiting to the different model states.
Modifying the transition probabilities is carried out by updating the elements of the matrix $\bm{A} \in \mathbb{R}^{K \times K}$.
The bounds of the element $\textbf{A}_{ij}$ are naturally $\mathcal{B}_{A_{ij}} = [0, 1]$ as $\bm{A}$ encodes probability measures.
We may skip a component $j$ if the adjacent element $A_{i,j+1}$ in the row $\bm{A}_i$ has a higher value when compared to the element $A_{ij}$.
Therefore, the algorithm modifies the transition matrix during optimization to get an optimal states sequence for the new task requirements.
To keep the sum of each row $\bm{A}_{i,:}$ equal to $1$, we normalize the matrix before using it in the retrieval phase.
The maximum number of parameters considered for optimization is $K(K - 1)/2$.
Each model state covers a specific region in space, and therefore changing the transition matrix may skip a region that was part of the demonstrated trajectories, which can be helpful to satisfy the specification $\varphi_2$.


\section{Experiments and Results}
We here show several experiments to test our approach on a variety of robot skills and using several STL specifications.
We evaluate our approach in simulation and on a real-robot task. 
In the former, we show how our approach can adjust the skill to changing task needs over time and space. 
In the latter, we test our approach on industrial assembly tasks to generate a robust trajectory that fulfils the task requirements.
In both settings, we use the Franka-Emika Panda $7$-DOF robotic arm.

\subsection{Simulation experiments}
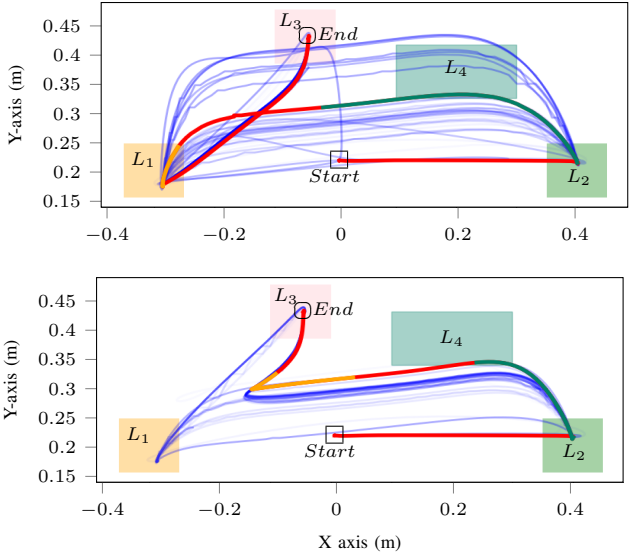
\begin{figure}[t]
    \centering
    \subfigure
    {
        \begin{tikzpicture}
            \begin{axis}[
                ylabel={Y-axis (m)},
                xtick distance=0.2,ytick distance=0.05,
                xtick pos=bottom, ytick pos=left,
                label style={font=\scriptsize},
                tick label style={font=\scriptsize},
                axis on top,
                width=7.0cm,
                scale only axis=true,
                enlargelimits=false,
                axis equal image,
                tick align=outside,
                ]
            \addplot graphics[xmin=-0.41, xmax=0.49, ymin=0.14, ymax=0.49,
                          includegraphics={width=8.5cm, trim={2.42cm 3.0cm 2.0cm 3.2cm},clip}] {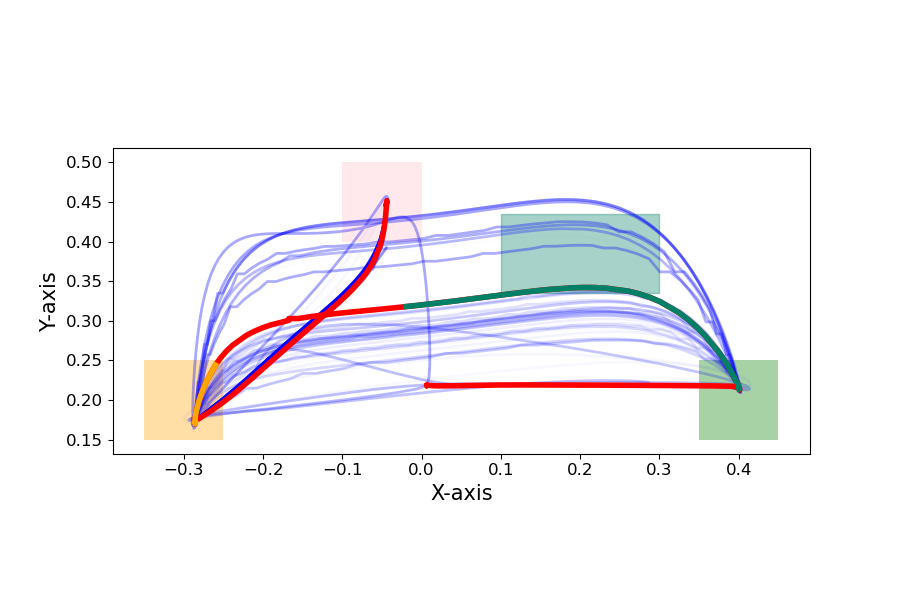};
            \end{axis}
        \node at (0.56, 0.63){\normalsize \scriptsize $L_1$};
        \node at (6.35, 0.40){\normalsize \scriptsize $L_2$};
        \node at (2.52, 2.475){\normalsize \scriptsize $L_3$};
        \node at (4.72, 1.85){\normalsize \scriptsize $L_4$};
        
        \node at (3.1, 0.415){\normalsize \scriptsize $Start$};
        \node at (3.14, 2.315){\normalsize \scriptsize $End$};
        
        \put(87,15){\framebox(6,6){}}
        \put(78,65){\oval(6,6){}}
        
        \end{tikzpicture}
        \label{fig:afterOptim1}
    }
    \\
    \subfigure
    {
        \begin{tikzpicture}
            \begin{axis}[
                xlabel={X axis (m)},ylabel={Y-axis (m)},
                xtick distance=0.2,ytick distance=0.05,
                xtick pos=bottom, ytick pos=left,
                label style={font=\scriptsize},
                tick label style={font=\scriptsize},
                axis on top, 
                width=7.0cm,
                scale only axis=true,
                enlargelimits=false,
                axis equal image,
                tick align=outside,
                ]
            \addplot graphics[xmin=-0.41, xmax=0.49, ymin=0.14, ymax=0.49,
                          includegraphics={width=8.5cm, trim={2.42cm 3.0cm 2.0cm 3.2cm},clip}] {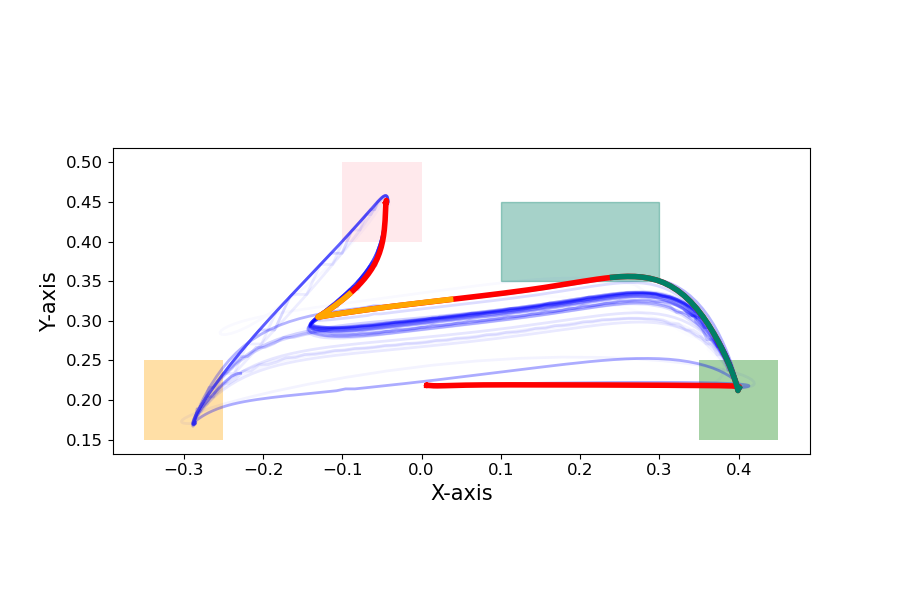};
            \end{axis}
        \node at (0.56, 0.63){\normalsize \scriptsize $L_1$};
        \node at (6.35, 0.40){\normalsize \scriptsize $L_2$};
        \node at (2.52, 2.475){\normalsize \scriptsize $L_3$};
        \node at (4.72, 1.92){\normalsize \scriptsize $L_4$};
        
        \node at (3.1, 0.415){\normalsize \scriptsize $Start$};
        \node at (3.14, 2.315){\normalsize \scriptsize $End$};

        \put(87,15){\framebox(6,6){}}
        \put(78,65){\oval(6,6){}}
        
        \end{tikzpicture}
        \label{fig:afterOptim2}
    }
    \vspace{-0.25cm}
    \caption{End-effector trajectories on $(x,y)$-plane (blue solid lines) at each BO iteration for $\varphi_1$ (top) and $\varphi_2$ (bottom).
    The color intensity depicts the optimization evolution and the red line represents the highest reward execution.
    The lines have colored patches representing the time interval when the end-effector has to reach the respective regions.}
    \label{fig:afterOptim}
    \vspace{-0.5cm}
\end{figure}

Our goal is to show that our approach handles spatial and temporal task constraints like staying or avoiding a region during specific time intervals.
We also show that it is possible to modify a skill by either adding new unseen goals or removing a part of the skill.
The number of iterations $N$ for these experiments is set to $32$.
Each iteration required around a minute to execute the skill on the simulator and less than a second to compute the rewards and obtain new parameters.
We recorded a couple of demonstrations as in Fig.~\ref{fig:afterOptim} on a real robot by moving the end-effector to three regions $L_{1:3}$ in task space in the same order.
For simplicity, we kept the initial and final positions of the end-effector unchanged in all the provided demonstrations.
Therefore, the problem is no more task parameterized as we do not observe the demonstrations from different perspectives, and hence the TP-HSMM model boils down to a simple HSMM model.
We train this model by setting $K=6$ experimentally.


        

For the STL specification $\varphi_1$ defined in~\eqref{Eqn: STL Spec 1}, to accommodate the task of visiting a new region and staying inside $L_2$ for several seconds, we adjusted the Gaussian component positions $\bm{\mu}_{1:6}$ and the duration probabilities $\bm{\mu}_{1:6}^S$, respectively.
Since this yields a total of $24$ parameters, which is a challenging dimensionality for BO, we further reduced the parameter domain by considering only the most relevant component positions for this task.
Specifically, we only optimized $\bm{\mu}_{1:2}$ as we observed that the addition of $L_4$ was required during the first few seconds of the trajectory.
Further, we used all the duration probabilities $\bm{\mu}_{1:6}^S$ as they all influence on obtaining the trajectory.
Finally, the reduced number of parameters is $12$.
Thus, the resulting parameter space is $\bm{\delta}_{\varphi_1} = \{ \bm{\mu}_{1:2}, \bm{\mu}_{1:6}^S\}$.

\begin{figure}[t]
    \centering
    \subfigure
    {
        \begin{tikzpicture}
            \begin{axis}[
                ylabel={Distance (m)},
                xtick distance=5,ytick distance=0.05,
                xtick pos=bottom, ytick pos=left,
                label style={font=\scriptsize},
                tick label style={font=\scriptsize, /pgf/number format/fixed},
                axis on top, 
                width=7.0cm,
                height=2.4cm,
                scale only axis=true,
                enlargelimits=false,
                tick align=outside,
                ]
            \addplot graphics[xmin=0, xmax=32, ymin=-0.1, ymax=0.185,
                          includegraphics={width=8.5cm, trim={2.9cm 1.32cm 2.55cm 1.0cm},clip}] {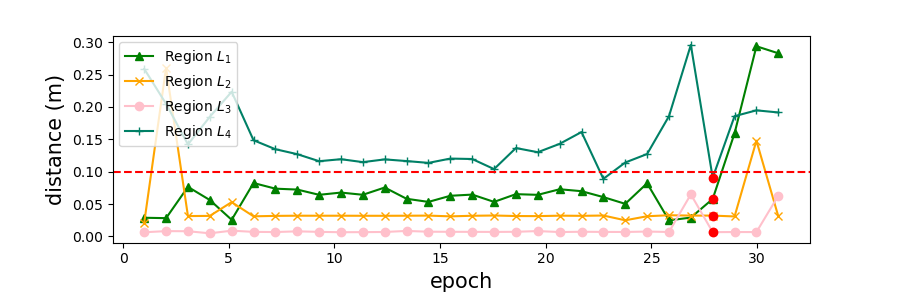};
            \end{axis}
        \end{tikzpicture}
    \label{fig:distToGoal1}
    }
    \\
    \subfigure
    {
        \begin{tikzpicture}
            \begin{axis}[
                xlabel={Epoch},ylabel={Distance (m)},
                xtick distance=5,ytick distance=0.05,
                xtick pos=bottom, ytick pos=left,
                label style={font=\scriptsize},
                tick label style={font=\scriptsize, /pgf/number format/fixed},
                axis on top, 
                width=7.0cm,
                height=2.4cm,
                scale only axis=true,
                enlargelimits=false,
                tick align=outside,
                ]
            \addplot graphics[xmin=0, xmax=32, ymin=-0.1, ymax=0.19,
                          includegraphics={width=8.5cm, trim={2.9cm 1.32cm 2.55cm 1.0cm},clip}] {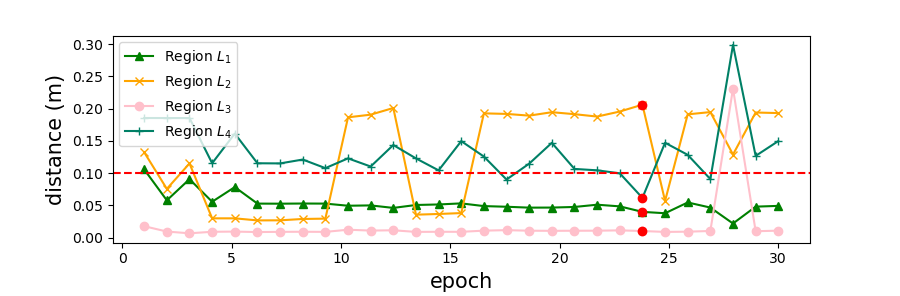};
            \end{axis}
        \end{tikzpicture}
    \label{fig:distToGoal2}
    }
    \vspace{-0.625cm}
    \caption{Signed distance function for each region under respective time constraints for $\varphi_1$ (top) and $\varphi_2$ (bottom).
    The horizontal dotted line shows the region's boundary in space.
    The positive and negative values denote the distances to the regions and boundaries when inside the regions respectively.
    The red dot depicts the iteration with the maximum reward.}
    \label{fig:distToGoal}
\end{figure}

\begin{figure}[t]
    \centering
    \includegraphics[width=5.5cm, trim={0.0cm 0.0cm 0.0cm 1.0cm},clip]{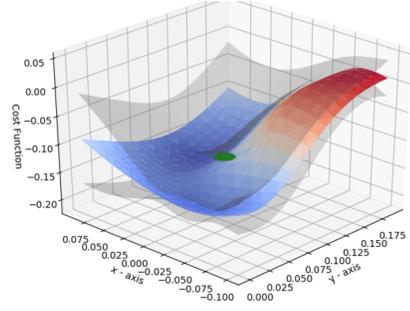}
    \caption{BO surrogate Gaussian Process model projected on the $(x, y)$ positions of component $3$ of the skill. The cost denotes negative STL Robustness. The coloured surface shows the mean of the model. The grey surface is the $\pm$ variance, and the green point is the location of the minimum underlying cost.}
    \label{fig:3dimGP}
    \vspace{-0.2cm}
\end{figure}

The trajectory of the end-effector in Fig.~\ref{fig:afterOptim}-top shows that the additional time constraint on $L_2$ to stay inside the region for the entire $5$ seconds, depicted by the yellow trajectory segment, is satisfied.
Moreover, the trace also shows the inclusion of $L_4$ into the trajectory.
We can observe in Fig.~\ref{fig:distToGoal}-top that all the regions were visited under the respective time bounds after several iterations.
Figure~\ref{fig:3dimGP} shows the low-dimensional BO surrogate model representation, i.e. the Gaussian Process of the third component position along $(x,y)$ axes, after evaluating $32$ iterations.
We can see that by translating the third component around $(0.025, 0.5)$ in Fig.~\ref{fig:3dimGP}, the cost is minimized (i.e. task satisfaction is maximized).

For the second task specification $\varphi_2$ , along with $\bm{\mu}_{1:2}$ and $\bm{\mu}_{1:6}^S$, we also considered the transition matrix $\textbf{A}$. 
This allows us to remove regions visited in initial demonstrations by pruning transitions to irrelevant Gaussian components.
For this task, we reduce the number of parameters by restricting the transition matrix to the entries
\begin{equation}
    \begingroup
        \setlength\arraycolsep{2pt}
        \textbf{A}_{red} = \begin{pmatrix}
            0 & A_{01} & A_{02} & 0         & 0         & 0          \\
            0 & 0         & A_{12} & A_{13} & 0         & 0          \\
            0 & 0         & 0         & A_{23} & A_{24} & 0          \\
            0 & 0         & 0         & 0         & A_{34} & A_{35} \\
            0 & 0         & 0         & 0         & 0         & A_{45} \\
            0 & 0         & 0         & 0         & 0         & 0
        \end{pmatrix},
    \endgroup
\end{equation}
effectively preventing to skip multiple components in immediate sequence to limit severe deviations from the original model.
Therefore, the number of parameters reduces to $21$ from the initial $27$ and the final parameter space is given by $\bm{\delta}_{\varphi_2} = \{ \bm{\mu}_{1:2}, \bm{\mu}_{1:6}^S, \bm{A}_{red} \}$.

\begin{figure}[t]
    \centering
    \subfigure
    {
        \begin{tikzpicture}
            \begin{axis}[
                xlabel={X axis (m)},ylabel={Cost Function},
                xtick distance=0.50,ytick distance=0.1,
                xtick pos=bottom, ytick pos=left,
                label style={font=\scriptsize},
                tick label style={font=\scriptsize},
                axis on top,
                width=4.6cm,
                height=4.0cm,
                scale only axis=false,
                enlargelimits=false,
                tick align=outside,
                ]
            \addplot graphics[xmin=-1.0, xmax=1.0, ymin=-0.20, ymax=0.20,
                          includegraphics={width=4.0cm, trim={2.4cm 1.7cm 2.1cm 1.6cm},clip}] {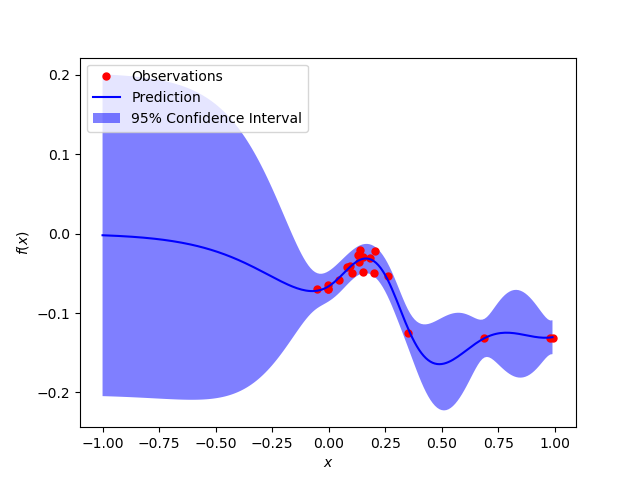};
            \end{axis}
        \end{tikzpicture}
        \label{fig:gpdp0}
    }
    \subfigure
    {
        \begin{tikzpicture}
            \begin{axis}[
                xlabel={X axis (m)},
                xtick distance=0.50,
                xtick pos=bottom,
                ytick = \empty,
                label style={font=\scriptsize},
                tick label style={font=\scriptsize},
                axis on top,
                width=4.6cm,
                height=4.0cm,
                scale only axis=false,
                enlargelimits=false,
                tick align=outside,
                ]
            \addplot graphics[xmin=-1.0, xmax=1.0, ymin=-0.20, ymax=0.20,
                          includegraphics={width=4.0cm, trim={2.4cm 1.7cm 2.1cm 1.6cm},clip}] {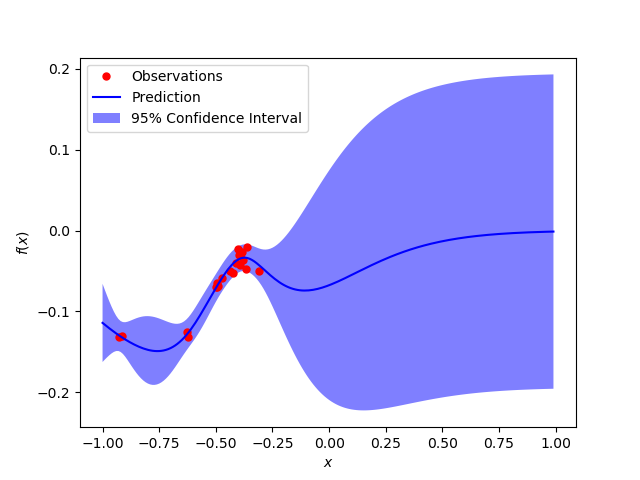};
            \end{axis}
        \end{tikzpicture}
        \label{fig:gpdp4}
    }
    \vspace{-0.25cm}
    \caption{Gaussian Process of the duration probabilities for component $1$ (left) and component $3$ (right) at the end of experiment for the test $\varphi_2$. The solid blue line represents the predictions, and the red dots depict the observations. The coloured area corresponds to the $95\%$ confidence interval.
    }
    \label{fig:2dGP}
    \vspace{-0.45cm}
\end{figure}

\begin{figure}[t]
    \centering
    \subfigure
    {
    \includegraphics[width=3.15cm, trim={0.0cm 0.0cm 0.0cm 0.0cm}, clip]{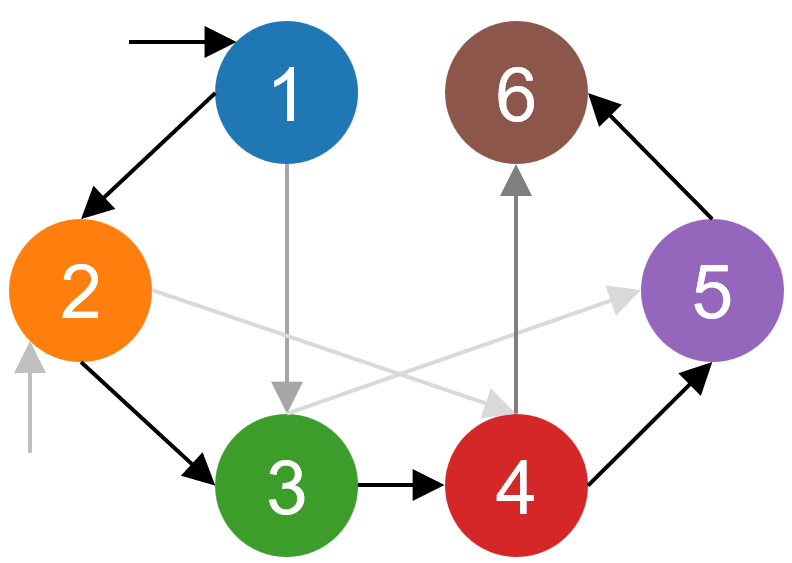}
    \label{fig:TransitionBefore}
    }
    \subfigure
    {
    \includegraphics[width=3.15cm, trim={0.0cm 0.0cm 0.0cm 0.0cm}, clip]{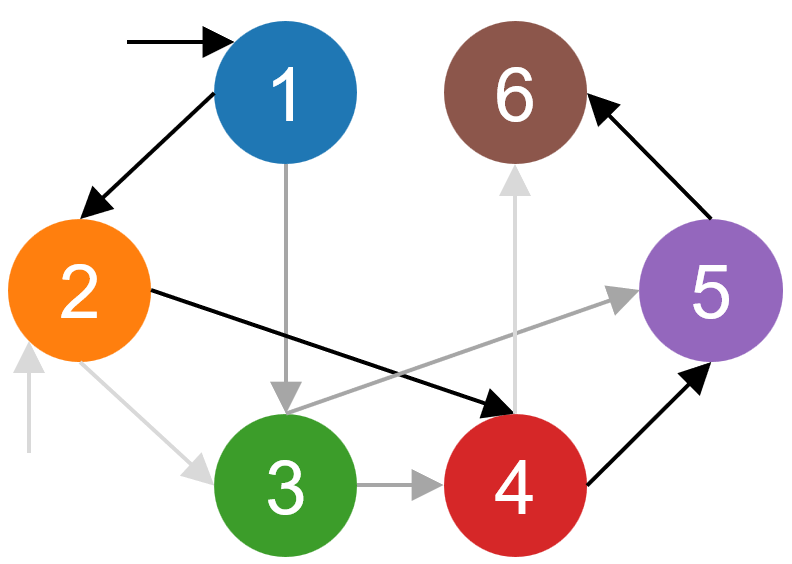}
    \label{fig:TransitionAfter}
    }
    \caption{Transition graph extracted from the transition matrix for the STL specification $\varphi_2$ before (left) and after (right) optimization. The black arrows depict the final transition sequence. Note that the components shown here do not associate to specific task requirements. The LfD model parameters are exclusively capturing the spatial and temporal patterns of the demonstrations, but later adapted according to new task requirements.}
    \label{fig:TransitionProb}
    \vspace{-0.45cm}
\end{figure}

After optimizing the reduced parameter set, Fig.~\ref{fig:afterOptim}-bottom depicts the trace of the end-effector at each iteration.
As shown in Fig.~\ref{fig:distToGoal}-bottom, the region $L_2$ is avoided and the region $L_4$ is added as specified in the STL constraint.
Figure~\ref{fig:2dGP} shows the GP for duration probability factors after $32$ iterations.
It shows that to minimize the cost, the duration probability $\mu_1^S$ has to be moved to the right, therefore increasing the time to stay in the Gaussian component $1$.
Similarly, $\mu_3^S$ has to be moved to the left, thus reducing the time to stay in the corresponding component.
Investigating the skill model more closely, the aforementioned results are reasonable as the component $3$ is located near $L_2$, so reducing $\mu_3^S$ and increasing $A_{24}$ transition probability will skip that region as depicted in Fig.~\ref{fig:TransitionProb}.

The foregoing results show that optimizing LfD parameters using STL specifications can accommodate temporal constraints like staying in the region $L_2$ for $5$ seconds. 
Similarly, satisfying spatial properties like visiting a new region $L_4$ by changing the trajectory can be achieved.

\subsection{Robot Experiment}
\begin{figure}[t]
    \centering
    \includegraphics[width=5.5cm, trim={0.0cm 0.0cm 0.0cm 0.0cm},clip]{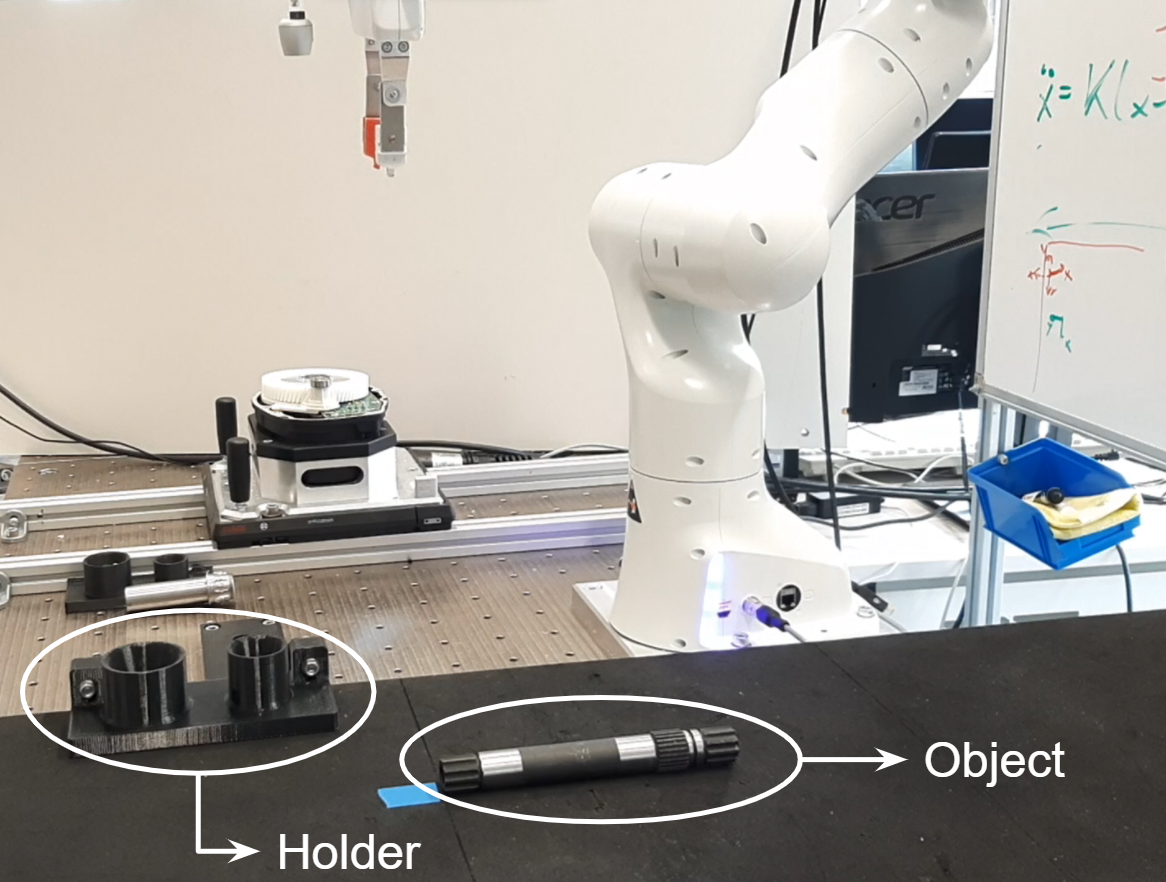}
    \caption{Robot setup with a $7$-DOF robot manipulator on an industrial assembly setup. The table has an object (a shaft) and a holder to place it.}
    \label{fig:RobotSetup}
    \vspace{-0.2cm}
\end{figure}

We designed a robot experiment to test our approach on an industrial assembly task as shown in Fig.~\ref{fig:RobotSetup}.
The robot has to pick up a shaft, re-orient it, and insert it into a specific location.
We show that our approach not only satisfies the task but also attempts to find the most robust solution.
We defined two different skills: One to pick up the object at a fixed position, and the other to re-orient and insert it into the desired location.
We trained two HSMM models (by setting $K=6$) with one demonstration for each skill.
Finally, we created a behaviour by sequentially combining the aforementioned skills with end-effector grasp and release actions.
The BO optimization was carried out for $16$ iterations.

\begin{figure}[t]
    \centering
    \subfigure
    {
    \includegraphics[width=4.0cm, trim={8.0cm 3.0cm 8.5cm 0.0cm},clip]{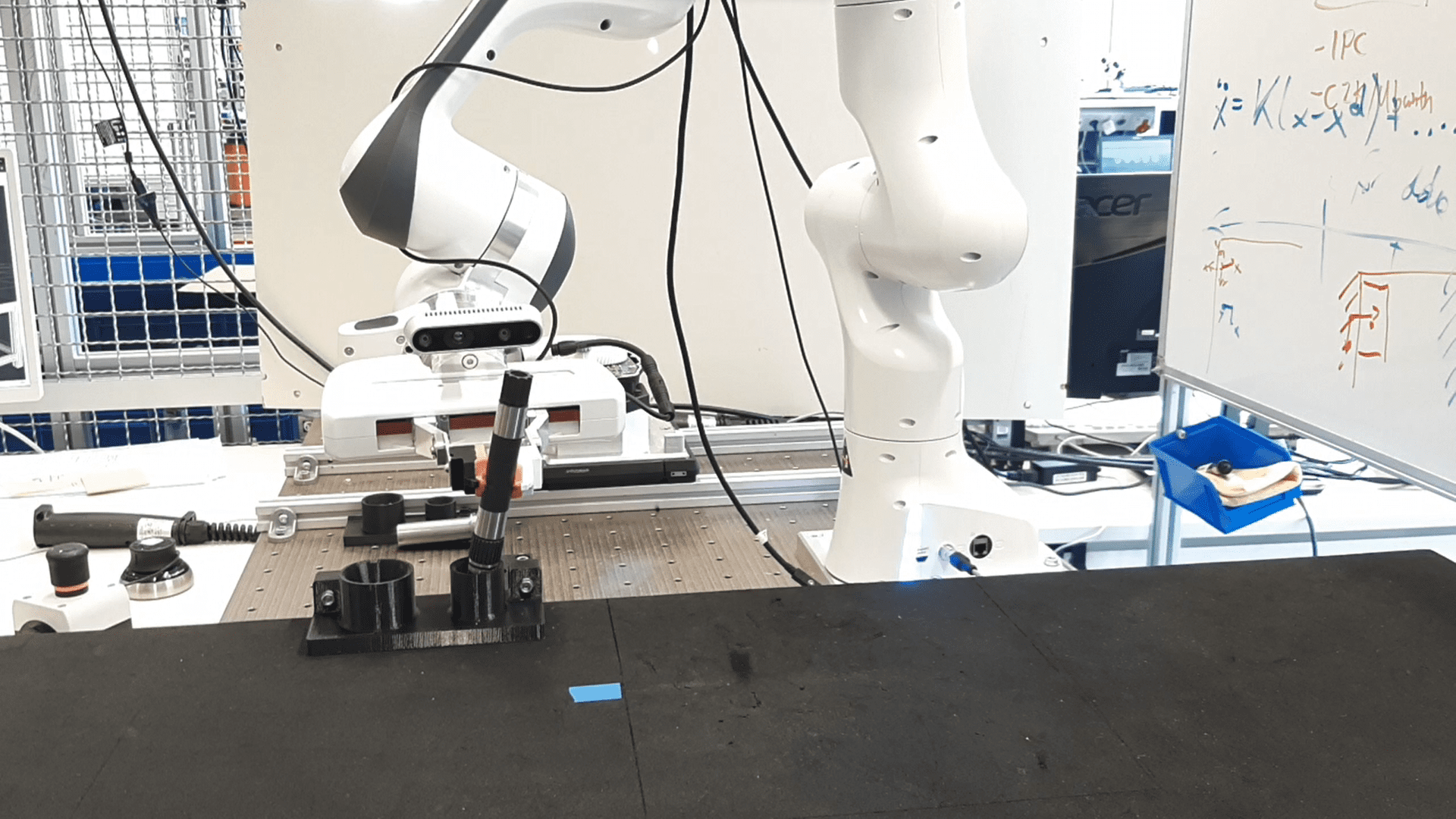}
    \label{fig:InsertionBefore}
    }
    \subfigure
    {
            \includegraphics[width=4.0cm, trim={8.0cm 2.5cm 7.0cm 0.0cm},clip]{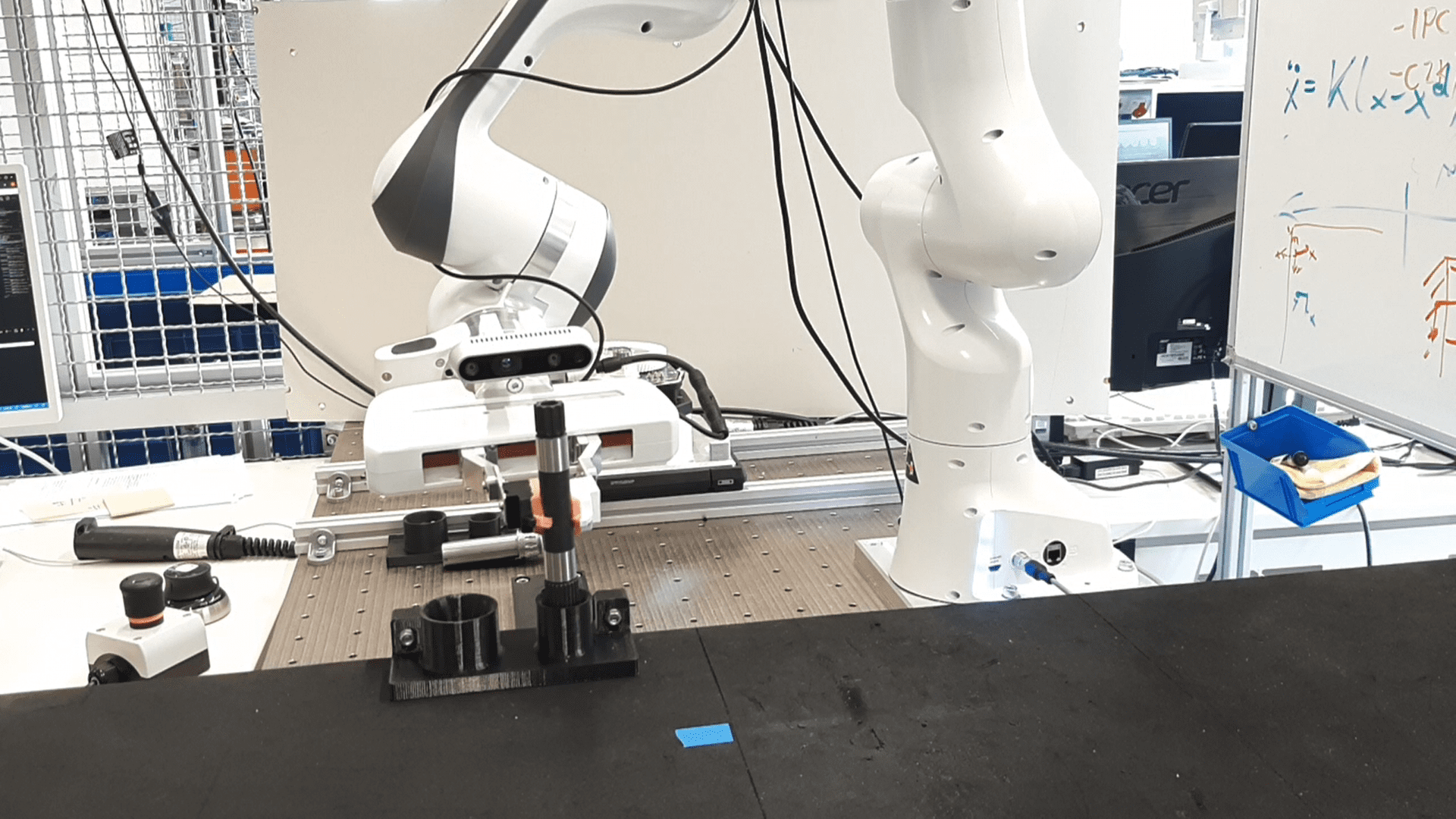}
            \label{fig:InsertionAfter}
    }
    \caption{Object insertion before (left) and after (right) optimization.}
    \label{fig:ObjectInsertion}
\end{figure}

Let us consider the task: Pick up the object within $20$ seconds and reduce the contact forces to less than $2$N.
The STL specification $\varphi_3$ for this task is,
\begin{equation}
\begin{aligned}
    \varphi_3 = & \mathbf{F}_{[0,20]} \varphi_{obj} \land \mathbf{F} \varphi_{force}, \\
    \varphi_{obj} \myeq & x_{obj,lb} \myless x \myless x_{obj,ub} \ \land \
                      y_{obj,lb} \myless y \myless y_{obj,ub} \ \land \\
                    & z_{obj,lb} \myless z \myless z_{obj,ub} , \\
    \varphi_{force} = & |F_{contact}| < 2.0.
\end{aligned}
\label{Eqn: STL PickInsert}
\end{equation}
The optimization parameters are the transition matrix $\textbf{A}_{red}$ and duration probabilities $\bm{\mu}_{1:6}^S$ of the picking skill and the component position $\mu_{6}$ of the insertion skill.
We considered the norm of the contact forces as another signal along with the end-effector trajectory.
We got access to contact forces using a $6$D force-torque sensor mounted at the robot end-effector.

\begin{figure}[t]
    \centering
        \begin{tikzpicture}
            \begin{axis}[
                xlabel={Execution Time (Sec)},ylabel={Contact Force (N)},
                xtick distance=10,ytick distance=5.0,
                xtick pos=bottom, ytick pos=left,
                label style={font=\scriptsize},
                tick label style={font=\scriptsize, /pgf/number format/fixed},
                axis on top, 
                width=7.25cm,
                height=1.8cm,
                scale only axis=true,
                enlargelimits=false,
                tick align=outside,
                ]
            \addplot graphics[xmin=-3.5, xmax=69.0, ymin=-1, ymax=20,
                          includegraphics={width=8.5cm, trim={3.15cm 1.1cm 2.50cm 1.0cm},clip}] {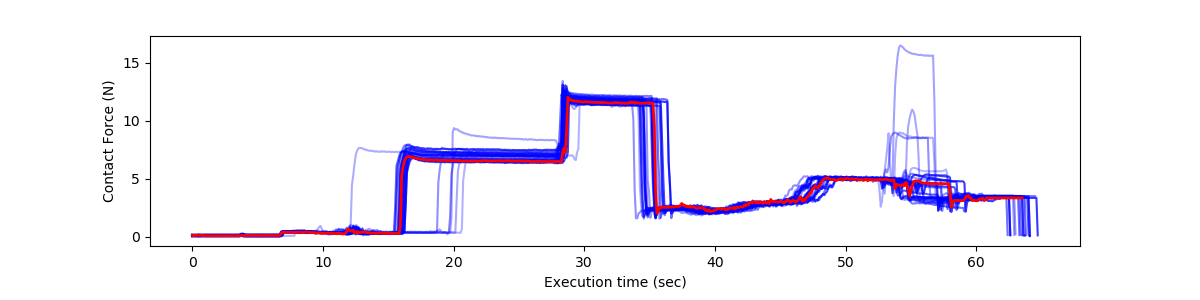};
            \end{axis}
        
        \node at (4.9, 1.40){\normalsize \scriptsize Time of};
        \node at (4.9, 1.24){\normalsize \scriptsize insertion};
        
        \put(170,27){\oval(28,42){}}
        
        \end{tikzpicture}
    \vspace{-0.4cm}
    \caption{Time series of the norm of the contact forces during the entire trajectory. The color intensity represents the optimization evolution. The circled time period ($52-58$ sec) corresponds to the image frames in Fig~\ref{fig:ObjectInsertion}.}
    \vspace{-0.35cm}
    \label{fig:robotExp2force}
\end{figure}
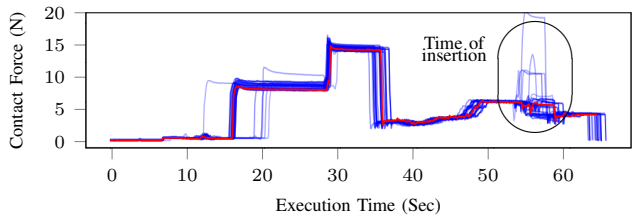

Figure~\ref{fig:ObjectInsertion}-left shows the initially learned LfD skill, which is sub-optimal due to the shift in the end-effector position and the holder.
However, insertion after refinement, as shown in Fig.~\ref{fig:ObjectInsertion}-right overcomes such shift so that the object does not hit the edges of the holder. 
This in turn reduces the contact forces as shown in Fig.~\ref{fig:robotExp2force}.
Note that there is no time constraint for $\varphi_{force}$ and in Fig.~\ref{fig:robotExp2force}.
The contact force is less than $2$N at the beginning, which means, the Eventually operator is already positive.
Albeit, the optimization finds a robust trajectory due to the property of robustness degree to guide towards task satisfaction.
Note that to achieve similar results without our approach, we may need a force-sensitive LfD framework to specifically consider the required force patterns during the demonstration phase of the insertion task.

\begin{figure}[t]
    \centering
        \begin{tikzpicture}
            \begin{axis}[
                xlabel={Epoch},ylabel={Execution Time (Sec)},
                xtick distance=1,ytick distance=20,
                xtick pos=bottom, ytick pos=left,
                label style={font=\scriptsize},
                tick label style={font=\scriptsize, /pgf/number format/fixed},
                axis on top, 
                width=7.25cm,
                height=1.8cm,
                scale only axis=true,
                enlargelimits=false,
                tick align=outside,
                ]
            \addplot graphics[xmin=0.4, xmax=16.5, ymin=0, ymax=89,
                          includegraphics={width=8.5cm, trim={2.0cm 1.5cm 3.6cm 0.8cm},clip}] {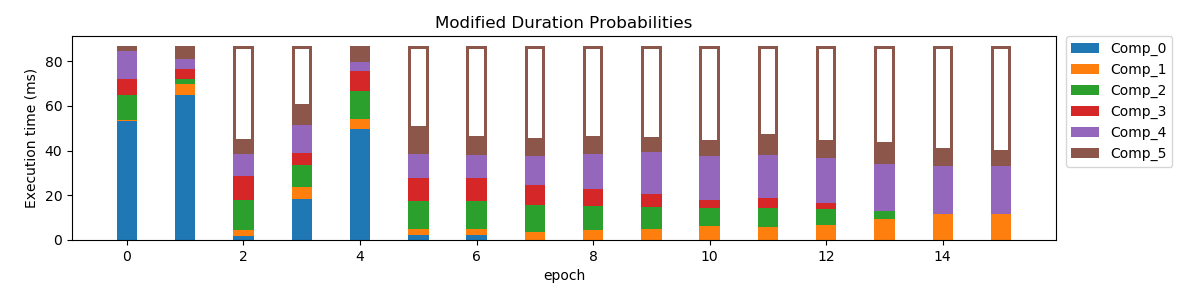};
            \end{axis}
        \end{tikzpicture}
    \vspace{-0.4cm}
    \caption{Stacked bar char depicting the duration mean at each iteration. Each color represents a component of the insertion skill. The $y$-axis shows the duration of each component's trajectory, and the white stack represents the reduced time.}
    \label{fig:robotExp2dp}
    \vspace{-0.35cm}
\end{figure}

Figure~\ref{fig:robotExp2dp} depicts the change in duration probabilities at each iteration.
It can be observed that the number of components approaches $K=3$ instead of $K=6$ during optimization, which means several components have been skipped due to the modifications of the transition matrix $\bm{A}$.
This can be interpreted as the skill being simple enough to be defined with fewer components.
Note that the approach does not modify the total duration of the skill but instead changes the duration to stay in each component.
Accordingly, the white bars in Fig.~\ref{fig:robotExp2dp} represent the time at the end during which the end-effector does not move anymore, and we can remove that time from the trajectory if needed.
Overall, this shows the capability of our optimization approach to ensure more reliable and faster satisfaction of the stated STL specification.

\section{Conclusion}
We presented an algorithm to include formal task requirements in an LfD model which are hard to specify implicitly by demonstration.
Instead, they are defined formally as an STL specification and then, the model parameters are optimized to accommodate the additional requirements.
Robot Experiments indicated that our approach of combining STL and BO could capture spatial and temporal constraints and find optimum trajectories for the task.
Some benefits of our approach are due to BO: we do not need to explicitly model the environment nor differentiable objectives to adapt the skill.
Regarding STL, it allows us to define a broad variety of spatial and temporal task requirements in a user-friendly manner. 
Future work will leverage nested STL operators and address two well-known problems in BO: the curse of dimensionality~\cite{brochu2010tutorial} and the geometry of the parameter space, the latter naturally arising in several robotic applications~\cite{Jaquier21:GaBOMatern}.

\bibliography{Bibliography}
\bibliographystyle{ieeetr}

\end{document}